\documentclass{bmvc2k}

\title{NeuDonatello:\\ Uncertainty-Aware Framework for\\ Accurate Neural SDF Learning}

\addauthor{Alvin Jinsung Choi$^*$}{alvinjinsung@utexas.edu}{1, 2}
\addauthor{Wanhee Kim}{gml78905@kaist.ac.kr}{3}
\addauthor{Taeyun Kim}{ktw1404@naverlabs.com}{2,4}
\addauthor{Dasol Hong}{ds.hong@kaist.ac.kr}{2}
\addauthor{Wooju Lee}{dnwn24@etri.re.kr}{2,5}
\addauthor{Hyun Myung}{hmyung@kaist.ac.kr}{2}

\addinstitution{
 Department of Computer Science\\
 The University of Texas at Austin\\
 Texas, USA
}
\addinstitution{
 School of Electrical Engineering\\
 Korea Advanced Institute of Science and Technology (KAIST)\\
 Daejeon, Republic of Korea
}
\addinstitution{
 Robotics Program\\
 Korea Advanced Institute of Science and Technology (KAIST)\\
 Daejeon, Republic of Korea
}
\addinstitution{
 Robotics Group\\
 NAVERLABS\\
 Seongnam, Republic of Korea
}
\addinstitution{
 Field Robotics Research Section\\
 ETRI\\
 Daejeon, Republic of Korea
}

\runninghead{Choi \etal}{NeuDonatello}

\def\eg{\emph{e.g}\bmvaOneDot}

\def\etal{\emph{et al}\bmvaOneDot}

\usepackage{bmvcabbrv}

\usepackage{graphicx}
\usepackage{booktabs}

\usepackage{amsfonts}
\usepackage{amsmath}
\usepackage{multirow}
\usepackage[accsupp]{axessibility}
\usepackage{pifont}
\usepackage{makecell}
\usepackage{tabularx}

\begin{document}

\maketitle

\begingroup
\renewcommand\thefootnote{\fnsymbol{footnote}}
\footnotetext[1]{Work done while at KAIST.}
\endgroup

\begin{abstract}
Neural surface reconstruction has emerged as a powerful paradigm for recovering high-quality 3D surfaces from multi-view images. However, recovering accurate geometry solely from RGB images remains challenging due to uncertainties arising from textureless regions, occlusions, and inherent scene ambiguities. Existing methods often overlook such uncertainties, leading to inaccurate estimates of the signed distance function (SDF). We introduce \textbf{NeuDonatello}, a novel framework that models and leverages SDF uncertainty to improve surface reconstruction. Central to our approach is to model spatially varying uncertainty using a Monte Carlo sampling strategy. Using this uncertainty, we develop an adaptive regularization that selectively strengthens geometric constraints where RGB supervision is unreliable, avoiding incorrect surface reconstruction. We further introduce an uncertainty-aware scale parameter for the SDF-to-density conversion. Conditioned on uncertainty, this design enables more accurate modeling of spatially varying densities. Extensive experiments demonstrate that \textbf{NeuDonatello} achieves state-of-the-art reconstruction accuracy, with robust performance across diverse scenes using only posed RGB images.
\end{abstract}    
\section{Introduction}
\label{sec:intro}

Reconstructing 3D surfaces from multi-view RGB images remains a central topic in computer vision, with applications spanning robotics, virtual reality, and computer graphics. Traditional methods, such as structure-from-motion~\cite{snavely2006photo, agarwal2011building, schonberger2016structure} and multi-view stereo~\cite{furukawa2009accurate, schonberger2016pixelwise, merrell2007real}, often struggle with completeness and smoothness in scenes with complex geometry or low texture. Neural implicit representations have emerged as a powerful alternative, offering improved quality through continuous scene models parameterized by neural networks. These approaches represent geometry as implicit functions, such as signed distance functions (SDFs)~\cite{yariv2020multiview, yariv2021volume, wang2021neus, huang2024sc}. Combined with differentiable volume rendering~\cite{mildenhall2020nerf}, they enable end-to-end optimization from posed RGB images. The continuity and smoothness of neural networks serve as a strong inductive bias, facilitating accurate 3D reconstruction even in challenging regions.


\begin{figure}[t!]
    \centering
        \centerline{\includegraphics[width=0.7\columnwidth]{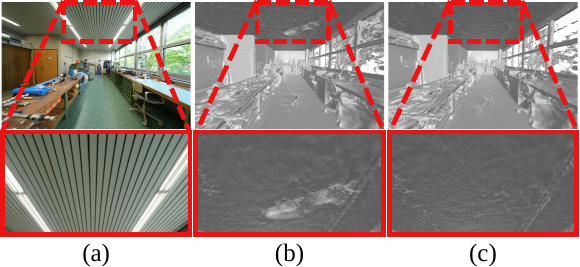}}
        \caption{(a) Region with lighting variation and limited views. (b) Without modeling uncertainty and using uniform regularization, the model produces artifacts and inaccurate geometry. (c) The proposed method leverages uncertainty to resolve ambiguity, producing accurate surface reconstruction.}
        \label{fig_1}
\end{figure}


Despite recent progress, recovering accurate surfaces remains challenging. A core difficulty arises from inherent ambiguity in multi-view surface reconstruction. Given a finite set of posed RGB images, multiple surfaces can produce similar renderings, making the inverse problem under-constrained. This ambiguity is particularly prominent in regions of high uncertainty, such as areas with textureless surfaces, lighting variations, or limited views. To address this, neural surface reconstruction methods jointly optimize photometric consistency and geometric constraints. Photometric loss encourages pixel-wise consistency between the predicted and observed images, while geometric regularization losses, such as eikonal constraints~\cite{gropp2020implicit} or smoothness terms~\cite{li2023neuralangelo, rosu2023permutosdf}, promote smooth and consistent surfaces. These constraints are important in regions where RGB supervision is unreliable. However, existing methods apply regularization uniformly across the entire scene, failing to account for spatially varying uncertainty and thus limiting their effectiveness in ambiguous regions.

In this paper, we introduce \textbf{NeuDonatello}, a high-fidelity neural surface reconstruction framework that models an implicit SDF representation through an uncertainty-aware pipeline. Our method estimates SDF uncertainty to distinguish geometrically ambiguous regions from well-constrained areas. Estimating this uncertainty is non-trivial due to the non-linear nature of SDF-based volume rendering, which hinders analytical uncertainty propagation. To address this, we propose a Monte Carlo sampling strategy to estimate SDF uncertainty directly from posed multi-view images. Leveraging this uncertainty, we develop an adaptive geometric regularization scheme that modulates the strength of regularization across space, allowing the model to apply stronger geometric constraints in regions where RGB supervision is unreliable. This enables accurate surface reconstruction even in the absence of auxiliary priors such as depth or normal maps. Furthermore, we introduce an uncertainty-aware SDF-to-density conversion to mitigate incorrect surface reconstruction. When the SDF prediction is unreliable, forcing a sharp boundary between empty and occupied space can lead to incorrect surfaces. Previous methods~\cite{wang2021neus, yariv2021volume, li2023neuralangelo, wang2024neurodin} only rely on global or position-dependent scale parameters, failing to capture spatial and directional ambiguity. To address this, we condition the scale parameter on position, viewing direction, and SDF uncertainty, allowing the model to account for local ambiguity and reduce density bias. Together, these components form an uncertainty-driven framework that robustly achieves high-fidelity surface reconstruction from RGB-only supervision, as shown in \cref{fig_1}.

We validate our approach through extensive experiments on the ScanNet++~\cite{yeshwanth2023scannet++} and Tanks and Temples~\cite{knapitsch2017tanks} datasets, covering diverse and challenging indoor and outdoor scenes. Our method consistently outperforms existing approaches in both quantitative metrics and visual fidelity. In particular, it achieves substantial improvements in geometrically ambiguous regions, such as textureless surfaces and sparsely observed areas, demonstrating the advantage of explicitly modeling and leveraging SDF uncertainty for robust RGB-based surface reconstruction.

In summary, our main contributions are as follows:

\begin{itemize}
\item We propose \textbf{NeuDonatello}, an uncertainty-aware neural surface reconstruction framework that estimates spatially varying SDF uncertainty via Monte Carlo sampling from posed multi-view RGB images.

\item We propose an adaptive geometric regularization scheme that modulates constraint strength according to local uncertainty, enabling accurate reconstruction in geometrically ambiguous regions without relying on auxiliary priors such as depth or normal maps.

\item We develop an uncertainty-aware SDF-to-density conversion that conditions the scale parameter on position, viewing direction, and SDF uncertainty, effectively mitigating density bias and improving surface fidelity.

\item We achieve state-of-the-art performance on challenging benchmarks, achieving consistent gains in both quantitative metrics and perceptual quality.
\end{itemize}

\section{Related Works}
\label{sec:related_works}

\subsection{Neural Surface Reconstruction}
NeRF~\cite{mildenhall2020nerf} introduced a framework for novel view synthesis using implicit neural representations and differentiable volume rendering. This paradigm has been extended to surface reconstruction by modeling geometry as an SDF and extracting surfaces via its zero-level set. Subsequent works have enhanced representational capacity using advanced positional encodings~\cite{rosu2023permutosdf, wang2023neus2}, incorporated auxiliary signals such as depth or normal priors~\cite{yu2022monosdf, wang2022neuris, darmon2022improving, fu2022geo, huang2024neusurf, wu2025sparis}, or proposed improved SDF-to-density conversions to address density bias~\cite{wang2021neus, zhang2023towards, xiao2024debsdf}. Neuralangelo~\cite{li2023neuralangelo} integrates multi-resolution hash encoding~\cite{muller2022instant} with coarse-to-fine optimization and numerical gradient estimation, while NeuRodin~\cite{wang2024neurodin} mitigates over-regularization via a two-stage training strategy. Despite these advances, prior methods rely on deterministic SDF predictions and apply geometric regularization uniformly across space, disregarding spatially varying uncertainty. In contrast, we explicitly model SDF uncertainty to guide surface learning, enabling robust and high-fidelity reconstruction even without auxiliary geometric priors such as depth or normal maps.

\subsection{Uncertainty in Neural Radiance Fields}
Recent works have incorporated uncertainty estimation~\cite{abdar2021review, kendall2017uncertainties} into neural radiance fields. NeRF-W~\cite{martinbrualla2020nerfw} models uncertainty to account for transient objects, lighting variation, and camera inconsistencies. Subsequent approaches~\cite{pan2022activenerf, jin2023neu, ran2023neurar} leverage uncertainty for active view selection. Other methods predict model uncertainty using variational inference~\cite{shen2021stochastic, shen2022conditional} or Laplace approximations~\cite{goli2024bayes}. However, these approaches primarily target radiance-based models for novel view synthesis and typically use uncertainty only for post-hoc analysis or inference-time confidence estimation rather than directly influencing the optimization. A few works~\cite{xiao2024debsdf, ni2024phyrecon} incorporate uncertainty into neural surface reconstruction, but they rely on strong geometric priors (\eg depth priors) and interpret uncertainty as noise in external signals to filter unreliable supervision. Despite incorporating uncertainty, they still oversmooth details and fail to preserve fine geometric structures. NeuDonatello fundamentally differs in that it models uncertainty as an intrinsic property of the RGB-only inverse rendering problem and estimates it directly in the 3D SDF representation. We further integrate this uncertainty into the optimization process in a closed-loop manner, enabling the model to adaptively resolve geometrically ambiguous regions while preserving fine details.

\begin{figure*}[t!]
    \centering
        \centerline{\includegraphics[width=0.95\linewidth]{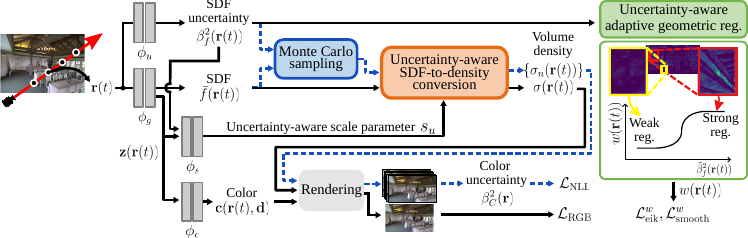}}
        \caption{The overview of NeuDonatello. The blue box and dotted lines (\cref{sec:3.2}) represent SDF uncertainty estimation via Monte Carlo sampling, supervised by the negative log-likelihood (NLL) loss. The green box (\cref{sec:3.3}) shows how uncertainty modulates regularization strength, which is stronger in uncertain regions (bright in the uncertainty map) and weaker in confident regions (dark), to improve geometric accuracy. The orange box (\cref{sec:3.4}) utilizes an uncertainty-aware scale parameter for the SDF-to-density conversion. (Best viewed in color.)}
        \label{fig:2}
\end{figure*}

\section{Proposed Method}
\label{sec:methods}

NeuDonatello reconstructs dense 3D geometry from multi-view images using an uncertainty-aware neural implicit framework. \cref{sec:3.1} reviews neural implicit surface reconstruction. \cref{sec:3.2} introduces our uncertainty modeling approach for estimating SDF uncertainty from multi-view RGB images. \cref{sec:3.3} describes how this uncertainty guides adaptive geometric regularization. \cref{sec:3.4} presents an uncertainty-aware scale parameter for SDF-to-density conversion that reduces bias and improves density accuracy. \cref{sec:3.5} outlines our optimization strategy. An overview of the pipeline is shown in \cref{fig:2}.

\subsection{Preliminaries} \label{sec:3.1}

SDF is widely used for implicit surface representation, where the surface is defined as the zero-level set. Integrating SDF representation into NeRF’s volume rendering framework~\cite{mildenhall2020nerf} has substantially improved reconstruction quality~\cite{wang2021neus, yariv2021volume}. Given a ray $\mathbf{r}=\{\mathbf{r}(t) = \mathbf{o} + t\mathbf{d} \mid t > 0\}$, 
where $\mathbf{o}$ is the camera origin and $\mathbf{d}$ is the viewing direction, a geometry network $\phi_g$ predicts the SDF value $f(\mathbf{r}(t))$ and geometric features $\mathbf{z}(\mathbf{r}(t))$ at sampled points. The SDF value is converted to volume density via a predefined function $\Phi_s$, typically the cumulative distribution function of a Laplace distribution: $\sigma(\mathbf{r}(t)) = \Phi_s(f(\mathbf{r}(t)))$. Here, $s$ is a scale parameter controlling the sharpness of the surface transition. The geometric features $\mathbf{z}(\mathbf{r}(t))$, along with the viewing direction $\mathbf{d}$ and surface normal $\mathbf{n}(\mathbf{r}(t)) = \nabla f(\mathbf{r}(t))$, are passed to a color network $\phi_c$ to predict the view-dependent radiance $\mathbf{c}(\mathbf{r}(t), \mathbf{d})$. The final pixel color $\hat{C}(\mathbf{r})$ is obtained via volume rendering as follows:

\begin{equation}
\label{eq:1}
    \hat{C}(\mathbf{r}) = \sum_{i=1}^{N} T(\mathbf{r}(t_i)) \alpha_i(\mathbf{r}(t_i))  \mathbf{c}(\mathbf{r}(t_i), \mathbf{d}),
\end{equation}
where $\alpha_i(\mathbf{r}(t_i)) = 1-\exp\left(-\sigma\left(\mathbf{r}(t_i)\right)\delta\left(\mathbf{r}(t_i)\right)\right)$ is the opacity of the $i$-th segment, $\delta(\mathbf{r}(t_i)) = \mathbf{r}(t_{i+1}) - \mathbf{r}(t_i)$ is the spacing, and $T(\mathbf{r}(t_i)) =  \prod_{j=1}^{i-1} (1 - \alpha_j(\mathbf{r}(t_j)))$ is the accumulated transmittance. Training is supervised using a photometric loss $\mathcal{L}_{\text{RGB}}$ between the rendered color $\hat{C}(\mathbf{r})$ and the ground-truth color $C(\mathbf{r})$ as follows:

\begin{equation} 
\label{eq:2}
    \mathcal{L}_{\text{RGB}} = \|\hat{C}(\mathbf{r}) - C(\mathbf{r})\|_1.
\end{equation}

To encourage geometric plausibility, regularization terms are typically applied to the predicted SDF representation. A common constraint is the eikonal loss~\cite{gropp2020implicit} which enforces that the gradient of the SDF has unit norm:

\begin{equation} 
\label{eq:3}
    \mathcal{L}_{\text{eik}} = \frac{1}{N} \sum_{i=1}^{N} \left(\|\nabla f(\mathbf{r}(t_i))\| - 1\right)^2.
\end{equation}

In addition, smoothness constraints~\cite{li2023neuralangelo, rosu2023permutosdf} are often applied to encourage local surface consistency. We adopt the smoothness loss from PermutoSDF~\cite{rosu2023permutosdf}:

\begin{equation} 
\label{eq:4}
    \mathcal{L}_{\text{smooth}} = \frac{1}{N} \sum_{i=1}^{N} \left(\mathbf{n}\left(\mathbf{r}(t_i)\right) \cdot \mathbf{n}\left(\mathbf{r}(t_i)+\epsilon\right) - 1\right)^2,
\end{equation}
where $\epsilon$ is a small spatial offset used to evaluate normal consistency between neighboring points.

However, uniformly applying such geometric regularization fails to account for spatial uncertainty, limiting its effectiveness in ambiguous regions where stronger geometric guidance is most needed.


\subsection{Uncertainty Modeling} \label{sec:3.2}

We explicitly model SDF uncertainty to distinguish between confident and uncertain regions in the learned representation. Mild regularization is sufficient in well-constrained areas where RGB supervision is reliable, as the photometric loss alone provides sufficient guidance. In contrast, highly uncertain regions benefit from stronger geometric regularization, which prevents convergence to implausible geometry and provides structural guidance.

To capture this spatial uncertainty efficiently, we adopt a Gaussian likelihood formulation~\cite{abdar2021review, kendall2017uncertainties}, modeling the SDF as a Gaussian distribution 

\begin{equation} 
\label{eq:5}
    \mathcal{N}\left(\bar{f}(\mathbf{r}(t)), \beta_f^2(\mathbf{r}(t))\right), 
\end{equation}
where $\bar{f}(\mathbf{r}(t)) \in \mathbb{R}$ and $\beta_f^2(\mathbf{r}(t)) \in \mathbb{R}^+$ denote the predicted mean and variance of the SDF, respectively. Both the geometry network $\phi_g$ and the uncertainty network $\phi_u$ are implemented as multi-layer perceptrons (MLPs). The geometry network outputs the mean SDF value and geometric features $\mathbf{z}(\mathbf{r}(t))$, while the uncertainty network predicts the SDF variance as follows:

\begin{equation} 
\label{eq:6}
    \begin{aligned}
    \{\bar{f}(\mathbf{r}(t)), \mathbf{z}(\mathbf{r}(t))\} &= \phi_g(\mathbf{r}(t))\\
    \text{and} \quad \beta_f^2(\mathbf{r}(t)) &= \phi_u(\mathbf{r}(t)).
    \end{aligned}
\end{equation}
This formulation captures uncertainty inherent in multi-view RGB reconstruction while remaining computationally tractable for dense 3D sampling. Unlike other uncertainty estimation methods such as variational inference~\cite{blundell2015weight, kingma2015variational} or deep ensembles~\cite{lakshminarayanan2017simple, zaidi2021neural}, our formulation performs a single forward pass per sampled point, making uncertainty estimation feasible during training. Consequently, the predicted uncertainty can be utilized in the optimization process as an active training signal that guides geometric refinement based on spatial reliability.

Since RGB supervision is applied in image space, we quantify the effect of SDF uncertainty on pixel colors by propagating it through the rendering process. This enables the network to receive gradients reflecting both rendered color accuracy and confidence in the underlying SDF predictions. In principle, uncertainty propagation can be performed analytically. However, the SDF-based volume rendering pipeline involves highly nonlinear components, making a closed-form solution intractable~\cite{lee2024bayesian}. To address this, we propose a simple yet effective Monte Carlo sampling strategy to propagate SDF uncertainty to the image plane. For each point along a ray, we draw $N_{\text{mc}}$ samples from the Gaussian distribution defined by the predicted SDF mean $\bar{f}(\mathbf{r}(t))$ and variance $\beta_f^2(\mathbf{r}(t))$:

\begin{equation} 
\label{eq:7}
    f_n(\mathbf{r}(t)) \overset{\text{i.i.d.}}{\sim} \mathcal{N}\left(\bar{f}(\mathbf{r}(t)), \beta_f^2(\mathbf{r}(t))\right),
\end{equation}
where $n=1, \cdots, N_{\text{mc}}$ and i.i.d. denotes that all samples are independent and identically distributed.  Each sampled SDF value is converted to volume density via $\sigma_n(\mathbf{r}(t)) = \Phi_s\left(f_n(\mathbf{r}(t))\right)$ where $s$ is a scale parameter. Together with the view-dependent radiance $\mathbf{c}_n(\mathbf{r}(t), \mathbf{d}) = \phi_c(\mathbf{z}(\mathbf{r}(t)), \mathbf{d}, \mathbf{n}_n(\mathbf{r}(t)))$, the resulting densities $\sigma_n(\mathbf{r}(t))$ are used in the volume rendering \cref{eq:1} to produce $N_{\text{mc}}$ pixel color estimates, denoted as $\{\hat{C}_n(\mathbf{r})\}_{n=1}^{N_{\text{mc}}}$. The rendered color uncertainty $\beta_C^2(\mathbf{r})$ is computed as the sample variance of these color predictions. This approach allows geometric uncertainty to be propagated into the pixel space, without requiring analytical derivatives through the rendering process.

We train the model using a negative log-likelihood (NLL) loss. Specifically, we adopt a stabilized variant~\cite{seitzer2022pitfalls}, which has been shown to improve optimization stability. Given the ground-truth pixel color $C(\mathbf{r})$ and the predicted color $\hat{C}(\textbf{r})$ computed from the SDF mean $\bar{f}(\mathbf{r}(t))$, the loss $\mathcal{L}_{\text{NLL}}$ is defined as follows:

\begin{equation} 
\label{eq:8}
    \mathcal{L}_{\text{NLL}} = \beta_C^{\gamma}(\mathbf{r}) \left( \frac{\| \hat{C}(\mathbf{r}) - C(\mathbf{r}) \|_1}{\beta_C(\mathbf{r}) + \varepsilon} + \log\left(\beta_C(\mathbf{r}) + \varepsilon\right) \right),
\end{equation}
where $\varepsilon$ is a small constant for numerical stability and $\gamma \in [0, 1]$ controls the weight of the uncertainty term. This loss formulation enables joint learning of both the predicted pixel color and its associated uncertainty $\beta_C^2(\mathbf{r})$. Crucially, it allows gradients to flow through both the SDF mean $\bar{f}(\mathbf{r}(t))$ and the SDF variance $\beta_f^2(\mathbf{r}(t))$, allowing end-to-end uncertainty learning from RGB supervision.


\subsection{Uncertainty-Aware Adaptive Geometric Regularization} \label{sec:3.3}

We adaptively scale geometric regularization strength based on the SDF uncertainty $\beta_f^2(\mathbf{r}(t))$, which is often ignored by existing methods. High uncertainty indicates unreliable RGB supervision, where stronger geometric regularization helps guide reconstruction toward plausible surface geometry. To enable this, we compute a normalized uncertainty $\tilde{\beta}_f^2(\mathbf{r}(t)) \in [0, 1]$ via min-max normalization across the current batch. We then define an adaptive weighting function $w(\mathbf{r}(t))$ to modulate regularization strength as follows:

\begin{equation} 
\label{eq:9}
    w(\mathbf{r}(t)) = \mu + \frac{\eta}{1 + \exp\left(-k \left(\tilde{\beta}_f^2(\mathbf{r}(t)) - \tau \right)\right)},
\end{equation}
where $\mu$ is a shift term, $\eta$ is a scale factor, $k$ controls sharpness, and $\tau$ sets the midpoint threshold. In this formulation, regularization strength increases in high uncertainty regions and decreases in well-constrained areas. We apply this adaptive weight to the eikonal loss $\mathcal{L}_{\text{eik}}^w$ and smoothness loss $\mathcal{L}_{\text{smooth}}^w$. The regularization losses are defined as:

\begin{align}
    \mathcal{L}_{\text{eik}}^w &= \frac{1}{N} \sum_{i=1}^{N} w(\mathbf{r}(t_i))\left(\|\nabla f(\mathbf{r}(t_i))\| - 1\right)^2, \label{eq:10} \\
    \mathcal{L}_{\text{smooth}}^w &= \frac{1}{N} \sum_{i=1}^{N} w(\mathbf{r}(t_i))\left(\mathbf{n}(\mathbf{r}(t_i)) \cdot \mathbf{n}(\mathbf{r}(t_i)+\epsilon) - 1\right)^2. \label{eq:11}
\end{align}

By adaptively weighting regularization based on SDF uncertainty, our approach applies stronger regularization in uncertain regions and weaker in confident regions. This strategy enforces geometric plausibility in ambiguous areas while preserving fine details where reliable cues are present.


\subsection{Uncertainty-Aware SDF-to-Density Conversion} \label{sec:3.4}

In the SDF-to-density conversion $\sigma(\mathbf{r}(t)) = \Phi_s(f(\mathbf{r}(t)))$, the scale parameter $s$ controls the sharpness of the density transition around the zero-level set of the SDF. A smaller $s$ yields a sharper and thinner density profile near the surface, while a larger $s$ produces a smoother and broader volumetric band. In ambiguous regions where SDF predictions are unreliable, a larger $s$ helps prevent incorrect surface localization, while a smaller $s$ is better suited for confident regions to capture sharp surface details. Prior works~\cite{wang2023adaptive, wang2024neurodin} introduced position-dependent scaling to adapt the density profile based on spatial variation. Although this improves flexibility, it does not account for view-dependent texture variation and fails to incorporate geometric uncertainty into the scaling process.

To address these limitations, we propose conditioning the scale parameter not only on position but also on SDF uncertainty and viewing direction. This design enables the network to adjust the density transition accordingly to both spatial and directional ambiguity, improving convergence and reconstruction accuracy, as illustrated in \cref{fig:3}. Specifically, the scale parameter is predicted by a scale network $\phi_s$, implemented as an MLP, as follows:

\begin{equation} 
\label{eq:12}
    s_u = \phi_s\left(\textbf{z}(\textbf{r}(t)), \textbf{d}, \beta_f^2(\mathbf{r}(t))\right).
\end{equation}

Conditioning on SDF uncertainty allows the scale parameter to reflect confidence in geometric predictions, promoting robust and accurate density modeling. Incorporating direction enables adjustment based on ray orientation, improving expressiveness in view-dependent regions.


\begin{figure}[t!]
    \centering
        \centerline{\includegraphics[width=0.5\columnwidth]{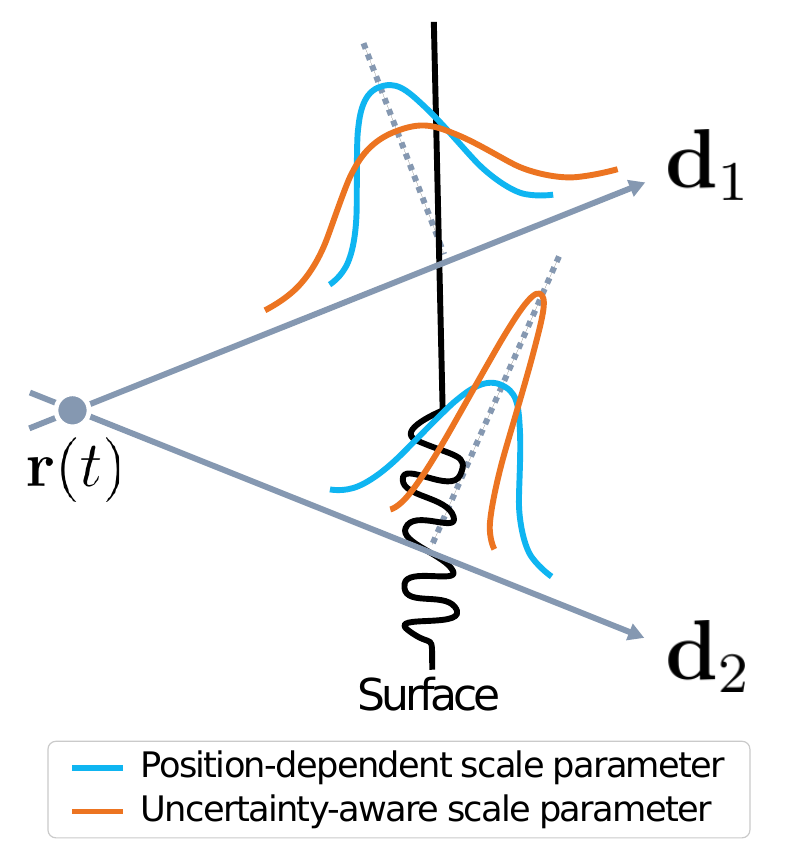}}
        \caption{Position-dependent scale parameter fails to model view-dependent density variations, leading to inaccurate reconstruction. Our uncertainty-aware scale parameter assigns high scale values in textureless regions (top) to prevent incorrect convergence and low scale values in textured regions (bottom) to enable accurate surface recovery.}
        \label{fig:3}
\end{figure}


\subsection{Optimization} \label{sec:3.5}

Our training framework adopts the two-stage optimization strategy proposed in NeuRodin~\cite{wang2024neurodin}. In the initial stage, we apply the NLL loss for uncertainty learning. We adopt the SDF-to-density conversion from VolSDF~\cite{yariv2021volume}, augmented with our uncertainty-aware scale parameter $s_u$. Additionally, we incorporate stochastic-step numerical gradient estimation $\hat{\nabla} f$ and explicit bias correction loss $\mathcal{L}_{\text{bias}}$ introduced in NeuRodin, both of which are effective in mitigating over-regularization. The resulting loss at this stage is:

\begin{equation} 
\label{eq:13}
    \mathcal{L}_{\text{init}} = \mathcal{L}_{\text{RGB}} + \lambda_{\text{NLL}}\mathcal{L}_{\text{NLL}} + \lambda_{\text{eik}}\mathcal{L}_{\text{eik}}(\hat{\nabla} f) + \lambda_{\text{bias}}\mathcal{L}_{\text{bias}}.
\end{equation}

In the refinement stage, we remove the NLL loss because the uncertainty has already been learned, and freeze the uncertainty network. Additionally, we remove the explicit bias correction and incorporate a smoothness constraint to promote local geometric consistency. For SDF-to-density conversion, we adopt the unbiased TUVR formulation~\cite{zhang2023towards} with the uncertainty-aware scale parameter $s_u$, which reduces bias and better preserves fine surface details. We further include the color regularization loss $\mathcal{L}_{\text{Lipschitz}}$ introduced in PermutoSDF~\cite{rosu2023permutosdf}. The overall loss function for the refinement stage is:

\begin{equation} 
\label{eq:14}
    \begin{aligned}
    \mathcal{L}_{\text{refine}} =\; & \mathcal{L}_{\text{RGB}} + \lambda_{\text{eik}}\mathcal{L}_{\text{eik}}^w(\nabla f) + \lambda_{\text{smooth}}\mathcal{L}_{\text{smooth}}^w \\
    & + \lambda_{\text{Lipschitz}}\mathcal{L}_{\text{Lipschitz}}.
    \end{aligned}
\end{equation}

For a detailed analysis of the impact of two-stage optimization, please refer to NeuRodin~\cite{wang2024neurodin}. Additional training details are provided in the Supplementary.

\section{Experiments}
\label{sec:experiments}

\noindent\textbf{Datasets.}
We evaluate our method on two standard benchmarks: ScanNet++~\cite{yeshwanth2023scannet++} and Tanks and Temples~\cite{knapitsch2017tanks}. ScanNet++ contains indoor scenes with frequent occlusions and textureless surfaces, resulting in high geometric uncertainty. Following NeuRodin~\cite{wang2024neurodin}, we report results on eight representative scenes. Tanks and Temples includes large-scale indoor and outdoor environments with diverse surface types, providing a challenging testbed. We evaluate diverse methods on six scenes from the training subset, following Neuralangelo~\cite{li2023neuralangelo}.

\par\medskip
\noindent\textbf{Baselines.}
For the ScanNet++ dataset, we compare our method with approaches that do not rely on external priors, including VolSDF~\cite{yariv2021volume}, Neuralangelo~\cite{li2023neuralangelo}, and NeuRodin~\cite{wang2024neurodin}, as well as MonoSDF~\cite{yu2022monosdf}, which incorporates monocular cues. For the Tanks and Temples dataset, we benchmark against COLMAP~\cite{schonberger2016structure} and recent neural methods such as NeuralWarp~\cite{darmon2022improving}, NeuS~\cite{wang2021neus}, Geo-NeuS~\cite{fu2022geo}, Neuralangelo~\cite{li2023neuralangelo}, and NeuRodin~\cite{wang2024neurodin}.

\par\medskip
\noindent\textbf{Metrics.}
We extract meshes using marching cubes at a fixed resolution of $2{,}048$, and evaluate reconstruction quality using six standard metrics: Accuracy, Completeness, Chamfer Distance, Precision, Recall, and F1-score.

\par\medskip
\noindent\textbf{Implementation Details.}
We use a multi-resolution hash grid for spatial encoding with 16 levels and resolutions from $2^5$ to $2^{11}$. Each entry stores an 8-dimensional feature vector, and each level contains up to $2^{19}$ entries. We set $N_{\text{mc}} = 10$. The loss weights are: $\lambda_{\text{NLL}} = 0.01$, $\lambda_{\text{eik}} = 0.01$, $\lambda_{\text{smooth}} = 0.005$, and $\lambda_{\text{Lipschitz}} = 10^{-5}$. $\lambda_{\text{bias}}$ is set to $0.2$ for outdoor scenes and linearly increased from $0.001$ to $0.1$ over the first $10{,}000$ iterations for indoor scenes. For the adaptive weighting function $w(\mathbf{r}(t))$, the shift $\mu$ is $0.8$, scale $\eta$ is $1.0$, sharpness $k$ is $10.0$, and midpoint $\tau$ is $0.5$. Additional details are provided in the Supplementary.


\subsection{ScanNet++}

We present quantitative results in \cref{tab:scannetpp_result} and qualitative comparisons in \cref{fig:4} on the ScanNet++ dataset. All methods are reproduced and evaluated on version 2, which differs from version 1 used in NeuRodin.\footnote{Version 1 is no longer available after the release of version 2.} NeuDonatello achieves state-of-the-art performance, outperforming prior RGB-only methods that do not model uncertainty and rely on uniform regularization without accounting for spatial ambiguity. We further compare with MonoSDF, a representative method incorporating monocular priors. NeuDonatello also surpasses MonoSDF, demonstrating that accurate geometry can be recovered without external geometric information by leveraging estimated uncertainty and jointly exploiting uncertainty-aware SDF-to-density conversion and adaptive geometric regularization.

\setlength{\tabcolsep}{0.7em}
\begin{table*}[t]
    \centering
    \begin{tabularx}{\textwidth}{@{}lrcccccc}
        \specialrule{.2em}{.1em}{.1em} 
        & & \multicolumn{6}{c}{Metric} \\ \cmidrule(l{.15em}r{.15em}){3-8}
        & \multicolumn{1}{c}{Method} 
        & \makecell{Acc. \\ ($\downarrow$)} 
        & \makecell{Comp. \\ ($\downarrow$)} 
        & \makecell{Pre. \\ ($\uparrow$)}
        & \makecell{Recall \\ ($\uparrow$)}
        & \makecell{Chamfer \\ ($\downarrow$)}
        & \makecell{F1-score \\ ($\uparrow$)} \\
        \cmidrule(l{0em}r{.15em}){2-2}
        \cmidrule(l{.15em}r{.15em}){3-8}
        & MonoSDF-MLP\textsuperscript{*}~\cite{yu2022monosdf} & \underline{0.053} & 0.052 & 0.542 & 0.546 & 0.053 & 0.542 \\
        & MonoSDF-Grid\textsuperscript{*}~\cite{yu2022monosdf} & 0.065 & \bfseries{0.040} & 0.579 & \underline{0.624} & \underline{0.052} & \underline{0.599} \\
        \cmidrule(l{0em}r{.15em}){2-2}
        \cmidrule(l{.15em}r{.15em}){3-8}
        & VolSDF~\cite{yariv2021volume} & 0.119 & 0.193 & 0.336 & 0.267 & 0.156 & 0.296  \\
        & Neuralangelo~\cite{li2023neuralangelo} & 0.156 & 0.092 & 0.492 & 0.557 & 0.124 & 0.534 \\
        & NeuRodin~\cite{wang2024neurodin} & 0.087 & 0.050 & \underline{0.592} & 0.606 & 0.069 & 0.596 \\
        & NeuDonatello (Ours) & \bfseries{0.052} & \underline{0.047} & \bfseries{0.618} & \bfseries{0.627} & \bfseries{0.049} & \bfseries{0.621} \\
        \specialrule{.2em}{.1em}{.1em}
    \end{tabularx}%
    \caption{
        Quantitative results on ScanNet++ dataset.
        Best result (bold). Second best result (underlined). \textsuperscript{*} indicates methods that use monocular depth and normal priors.
    }
    \label{tab:scannetpp_result}
\end{table*}

\begin{figure*}[t]
    \centering
    \centerline{\includegraphics[width=\linewidth]{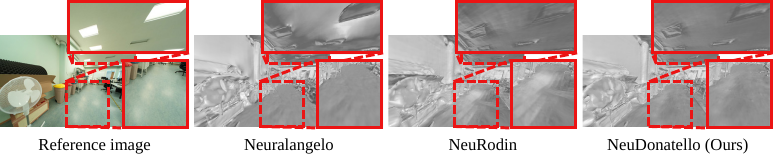}}
    \caption{
        Qualitative results on ScanNet++ dataset.
    }
    \label{fig:4}
\end{figure*}

Our uncertainty-aware modules reduce artifacts and structural degradation in geometrically ambiguous regions, including ceilings and floors. These areas suffer from unreliable RGB supervision due to low texture or strong reflections, making reconstruction particularly challenging. In \cref{fig:4}, the textureless ceiling and reflective floor cause failure cases in other methods: Neuralangelo produces a hole in the ceiling, while NeuRodin exhibits floor collapse. In contrast, NeuDonatello preserves surface continuity and accurately reconstructs geometry by leveraging SDF uncertainty to guide regularization and density modeling.


\setlength{\tabcolsep}{0.22em}
\begin{table*}[t]
    \centering
    \begin{tabularx}{\textwidth}{@{}lrccccccc}
        \specialrule{.2em}{.1em}{.1em}
        & & \multicolumn{6}{c}{Scene} \\ \cmidrule(l{.2em}r{.2em}){3-8}
        & \multicolumn{1}{c}{Method} & Barn & Caterpillar & Courthouse & Ignatius & Meetingroom & Truck & Mean \\
        \cmidrule(l{0em}r{.2em}){2-2}
        \cmidrule(l{.2em}r{.2em}){3-8}
        \cmidrule(l{.2em}r{0.2em}){9-9}
        & COLMAP~\cite{schonberger2016structure} & 0.55 & 0.01 & 0.11 & 0.22 & 0.19 & 0.19 & 0.21 \\
        & NeuS~\cite{wang2021neus} & 0.29 & 0.29 & 0.17 & 0.83 & 0.24 & 0.45 & 0.38 \\
        & NeuralWarp~\cite{darmon2022improving} & 0.22 & 0.18 & 0.08 & 0.02 & 0.08 & 0.35 & 0.15 \\
        & Geo-NeuS~\cite{fu2022geo} & 0.33 & 0.26 & 0.12 & 0.72 & 0.20 & 0.45 & 0.35 \\
        & Neuralangelo~\cite{li2023neuralangelo} & \underline{0.70} & \underline{0.36} & \bfseries{0.28} & \bfseries{0.89} & 0.32 & \bfseries{0.48} & \underline{0.50} \\
        & NeuRodin~\cite{wang2024neurodin} & \underline{0.70} & \underline{0.36} & 0.21 & \underline{0.87} & \underline{0.43} & \underline{0.47} & \bfseries{0.51} \\
        & NeuDonatello (Ours) & \bfseries{0.71} & \bfseries{0.37} & \underline{0.22} & 0.85 & \bfseries{0.44} & \bfseries{0.48} & \bfseries{0.51} \\
        \specialrule{.2em}{.1em}{.1em}
    \end{tabularx}%
    \caption{
        Quantitative results (F1-score ($\uparrow$)) on Tanks and Temples dataset. Best result (bold). Second best result (underlined).
    }
    \label{tab:tnt_result}
\end{table*}


\begin{figure*}[t]
    \centering
        \centerline{\includegraphics[width=\linewidth]{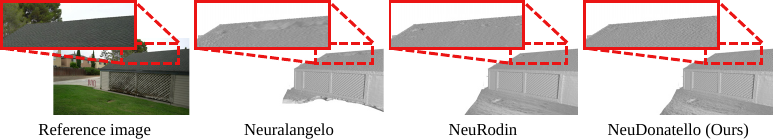}}
        \caption{
            Qualitative results on Tanks and Temples dataset.
        }
        \label{fig:5}
\end{figure*}

\subsection{Tanks and Temples}

We show quantitative results in \cref{tab:tnt_result} and qualitative comparisons in \cref{fig:5} on the Tanks and Temples dataset. NeuDonatello achieves the highest mean F1-score and consistently outperforms prior methods across both indoor and outdoor scenes. It demonstrates strong performance particularly in challenging cases with sparse views and textureless surfaces, such as Barn and Meetingroom. The adaptive geometric regularization helps preserve plausible structure in poorly observed regions, while the uncertainty-aware scale parameter maintains fine details and prevents convergence to erroneous surfaces.

Qualitative comparisons further highlight the benefits of our uncertainty-aware components under challenging conditions. In \cref{fig:5}, the roof region is sparsely observed due to limited camera coverage, making it difficult to reconstruct accurately. As a result, Neuralangelo and NeuRodin exhibit surface collapse or noticeable geometric deformation in this area. In contrast, NeuDonatello accurately preserves the overall structure and surface continuity by effectively leveraging the predicted SDF uncertainty.

\begin{figure}[t]
    \centering
        \centerline{\includegraphics[width=0.6\columnwidth]{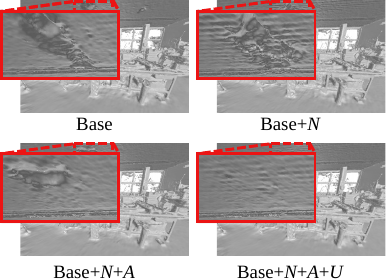}}
        \caption{Ablation results. \textit{N}: adds NLL loss for uncertainty learning. \textit{A}: applies adaptive geometric regularization. \textit{U}: incorporates uncertainty-aware scale parameter.}
        \label{fig:6}
\end{figure}


\setlength{\tabcolsep}{0.35em}
\begin{table}[t]
    \centering
    \begin{tabularx}{\columnwidth}{ccc|cccccc}
        \specialrule{.2em}{.1em}{.1em}
        \makecell{NLL \\ loss \\ (\textit{N})}
        & \makecell{Adaptive \\ geo. reg. \\ (\textit{A})}
        & \makecell{Uncertainty-aware \\ scale param. \\ (\textit{U})}
        & \makecell{Acc. \\ ($\downarrow$)}
        & \makecell{Comp. \\ ($\downarrow$)}
        & \makecell{Pre. \\ ($\uparrow$)}
        & \makecell{Recall \\ ($\uparrow$)}
        & \makecell{Chamfer \\ ($\downarrow$)}
        & \makecell{F1-Score \\ ($\uparrow$)} \\ 
        \cmidrule(l{0.33em}r{.33em}){1-9}
        & & & 0.088 & 0.048 & 0.580 & 0.602 & 0.068 & 0.591 \\
        \checkmark & & & 0.081 & 0.049 & 0.583 & 0.603 & 0.065 & 0.593 \\
        \checkmark & \checkmark & & 0.073 & 0.048 & 0.598 & 0.619 & 0.061 & 0.607 \\
        \checkmark & \checkmark & \checkmark & \bfseries{0.052} & \bfseries{0.047} & \bfseries{0.618} & \bfseries{0.627} & \bfseries{0.050} & \bfseries{0.621} \\
        \specialrule{.2em}{.1em}{.1em}
    \end{tabularx}%
    \caption{
        Ablation results of uncertainty-aware components evaluated on the ScanNet++ dataset.
    }
    \label{tab:ablation}
\end{table}


\setlength{\tabcolsep}{0.32em}
\begin{table}[t!]
    \centering
    \begin{tabularx}{\columnwidth}{ccc|cccccc}
        \specialrule{.2em}{.1em}{.1em}
        Position & Direction & SDF uncertainty
        & \makecell{Acc. \\ ($\downarrow$)}
        & \makecell{Comp. \\ ($\downarrow$)}
        & \makecell{Pre. \\ ($\uparrow$)}
        & \makecell{Recall \\ ($\uparrow$)}
        & \makecell{Chamfer \\ ($\downarrow$)}
        & \makecell{F1-Score \\ ($\uparrow$)} \\
        \cmidrule(l{0.33em}r{0.33em}){1-9}
        \checkmark & & & 0.073 & 0.048 & 0.598 & 0.619 & 0.061 & 0.607 \\
        \checkmark & \checkmark & & 0.054 & \bfseries{0.047} & 0.614 & 0.622 & 0.051 & 0.616 \\
        \checkmark & \checkmark & \checkmark & \bfseries{0.052} & \bfseries{0.047} & \bfseries{0.618} & \bfseries{0.627} & \bfseries{0.050} & \bfseries{0.621} \\
        \specialrule{.2em}{.1em}{.1em}
    \end{tabularx}
    \caption{
        Ablation results of scale parameter conditioning evaluated on the ScanNet++ dataset.
    }
    \label{tab:scale_param_ablation}
\end{table}


\subsection{Ablations}

We conduct an ablation study to analyze the contribution of each uncertainty-aware component in our framework on the ScanNet++ dataset. As shown in \cref{fig:6}, starting from the base model, introducing the NLL loss alone does not directly enhance surface quality and produces artifacts in highly uncertain regions. Nevertheless, it enables SDF uncertainty estimation and lays the foundation for downstream modules that explicitly leverage uncertainty during training. Applying adaptive geometric regularization improves surface fidelity by enforcing stronger constraints in ambiguous regions while preserving fine details in well-constrained areas. Finally, incorporating the uncertainty-aware scale parameter further refines surface reconstruction by reducing density bias near the zero-level set. \cref{tab:ablation} quantitatively confirms these observations, showing a consistent improvement in various metrics and the overall reconstruction quality as each component is integrated into the framework.

We also evaluate the impact of conditioning the scale parameter on SDF uncertainty. \cref{tab:scale_param_ablation} reports the quantitative results on the ScanNet++ dataset when the scale parameter is conditioned on position, direction, and SDF uncertainty, compared with conditioning only on position and direction or on position alone. Incorporating SDF uncertainty leads to an improvement in reconstruction performance, as it provides explicit information about local geometric ambiguity. This allows the model to predict scale values that better reflect the confidence of the SDF estimates, resulting in a more accurate density modeling and more stable surface reconstruction.


\begin{figure}[t]
    \centering
        \centerline{\includegraphics[width=1.0\columnwidth]{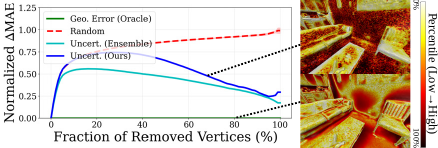}}
        \caption{Sparsification curve evaluated on the ScanNet++ dataset.}
        \label{fig:7}
\end{figure}


\setlength{\tabcolsep}{0.75em}
\begin{table}[t]
    \centering
    \begin{tabularx}{0.8\columnwidth}{c|ccc}
        \specialrule{.2em}{.1em}{.1em}
        Method & Random & Ensemble & NeuDonatello (Ours) \\
        \cmidrule(l{0.33em}r{0.33em}){1-4}
        Mean AUSE ($\downarrow$) & 0.721 & 0.376 & 0.458 \\
        \specialrule{.2em}{.1em}{.1em}
    \end{tabularx}
    \caption{
        AUSE results evaluated on the ScanNet++ dataset.
    }
    \label{tab:ause}
\end{table}


\subsection{Analysis on Uncertainty Estimation}

To validate that our uncertainty module captures intrinsic geometric ambiguity from RGB images alone, we analyze the relationship between predicted uncertainty and true geometric error. Given a reconstructed mesh $\mathcal{M}$ and ground truth $\mathcal{M}^{\text{gt}}$, we compute for each vertex $v_i \in \mathcal{M}$ its SDF uncertainty $\beta_f^2(v_i)$ and geometric error $e(v_i)=\|v_i-\Pi_{\mathcal{M}^{\text{gt}}}(v_i)\|_2$, where $\Pi_{\mathcal{M}^\text{gt}}(\cdot)$ denotes the closest-point projection onto the $\mathcal{M}^{\text{gt}}$. We perform a sparsification analysis by removing vertices based on either geometric error (oracle) or predicted uncertainty, and compute the mean absolute error (MAE) of the remaining vertices. We then report the MAE gap relative to the oracle ($\Delta$MAE) as well as the area under the sparsification error curve (AUSE).

The analysis is conducted directly in 3D, consistent with the uncertainty representation. We compare against random removal and an ensemble-based baseline, as no prior work estimates uncertainty in the 3D SDF representation. As shown in \cref{fig:7}, uncertainty-guided sparsification outperforms random removal and achieves performance comparable to the ensemble baseline, while being substantially faster. The correspondence between uncertainty and error, together with low AUSE values (\cref{tab:ause}), confirms that our method correctly identifies ambiguous regions in RGB-only surface reconstruction.



\subsection{Analysis on Efficiency}

We report computational cost in \cref{tab:comp_cost}. We compare against our baseline, which removes all uncertainty-aware modules. This baseline has identical computational cost to NeuRodin~\cite{wang2024neurodin}. We further compare against an ensemble-based baseline to evaluate the computational overhead of alternative uncertainty estimation approaches.

Our Monte Carlo sampling is implemented in a GPU-parallel manner, introducing only modest overhead relative to the baseline while remaining substantially more efficient than the ensemble-based method. This design remains computationally tractable even under dense 3D ray sampling, as stochastic evaluations are fully parallelized and does not require repeated full-network forward passes. These results demonstrate that our uncertainty modeling achieves strong performance gains with minimal additional computational cost.


    
    
    

\setlength{\tabcolsep}{1.0em}
\begin{table}[t!]
    \centering
    \begin{tabularx}{0.8\columnwidth}{c|ccc}
        \specialrule{.2em}{.1em}{.1em}
        Method & Baseline & Ensemble & NeuDonatello (Ours) \\
        \cmidrule(l{0.33em}r{0.33em}){1-4}
        Time/iter (s) & 0.11 & 0.82 & 0.14 \\
        VRAM (GB) & 5.4 & 16.2 & 7.9 \\
        \specialrule{.2em}{.1em}{.1em}
    \end{tabularx}
    \caption{
        Computational cost analysis.
    }
    \label{tab:comp_cost}
\end{table}
\section{Conclusion}
\label{sec:conclusion}


We propose NeuDonatello, an uncertainty-aware neural surface reconstruction framework that estimates SDF uncertainty from posed multi-view images and leverages it to guide accurate geometry reconstruction. Through Monte Carlo sampling, we identify geometrically ambiguous regions and adaptively modulate geometric regularization strength. We further introduce an uncertainty-aware SDF-to-density conversion by conditioning the scale parameter on position, direction, and uncertainty, thereby reducing bias and improving surface accuracy. Experiments on ScanNet++ and Tanks and Temples demonstrate that NeuDonatello outperforms prior methods, particularly in challenging regions with low texture or sparse views, highlighting the importance of uncertainty modeling in neural reconstruction. In future work, we plan to extend our framework to handle non-posed inputs and dynamic scenes for broader real-world applicability.

\bibliography{egbib}

\newpage
\setcounter{section}{0} 
\setcounter{table}{0} 
\setcounter{figure}{0} 

\renewcommand\thesection{\Alph{section}}
\renewcommand{\thetable}{\Alph{table}}
\renewcommand{\thefigure}{\Alph{figure}}

\section{Additional Implementation Details}

\subsection{Architecture}

NeuDonatello employs multi-resolution hash grid for spatial encoding with 16 levels and resolutions ranging from $2^5$ to $2^{11}$. Each hash entry stores an 8-dimensional feature vector, and each level allows up to $2^{19}$ entries. The geometry network $\phi_g$, the uncertainty network $\phi_u$, the color network $\phi_c$, and the scale network $\phi_s$ are all implemented as multi-layer perceptrons (MLPs). Specifically, $\phi_g$ has 1 hidden layer with 256 dimensions, $\phi_u$ has 2 hidden layers with 256 dimensions, and $\phi_c$ has 4 hidden layers with 256 dimensions each. The scale network $\phi_s$ is a single linear layer that outputs a scalar scale value.

\subsection{Training Details}

NeuDonatello integrates a proposal network following the design of Mip-NeRF 360~\cite{barron2022mip}, based on a compact hash grid representation. To account for appearance variation, we incorporate a learned appearance embedding similar to NeRF-W~\cite{martinbrualla2020nerfw}. For outdoor scenes, we additionally model the background using a dedicated network with a separate hash grid, inspired by techniques from NeRF++~\cite{zhang2020nerf++}.

During training, we sample 512 pixels per iteration and downsample the images by a factor of $2$ for the ScanNet++ dataset~\cite{yeshwanth2023scannet++}. For the Tanks and Temples dataset~\cite{knapitsch2017tanks}, we begin with 1024 pixel samples in the initial stage and increase to 8192 pixels in the refinement stage to improve reconstruction fidelity.

Our method is implemented in PyTorch and optimized using the Adam optimizer. The training is done for a total of $300{,}000$ iterations. We use a base learning rate of $0.001$ for both the neural networks and the hash grid, along with a weight decay of $0.01$. For the foreground model, the learning rate is decayed by a factor of 10 at $160{,}000$ and $240{,}000$ iterations. The background model is trained with an initial learning rate of $0.01$, which is gradually reduced to $0.0001$ using an exponential decay schedule. The proposal network uses the same initial rate of $0.001$, and its learning rate is decreased by a factor of 3 at steps $150{,}000$, $225{,}000$, and $270{,}000$. All experiments were conducted on a single NVIDIA A5000 GPU with 24GB of memory.


\subsection{Implementation Aspects}

\subsubsection{SDF-to-Density Conversion.}
For completeness, we summarize the two SDF-to-density conversion functions used in our two-stage training pipeline. In the initial stage, we adopt the SDF-to-density mapping from VolSDF~\cite{yariv2021volume}, modified with our uncertainty-aware scale parameter $s_u$ as follows:

\begin{equation} 
\label{eq:supp_1}
    \begin{aligned}
        \sigma(\mathbf{r}(t)) &= \Phi_{s_u}^{\text{VolSDF}}(f(\mathbf{r}(t))) \\
        =&
        \begin{cases}
        \frac{1}{2s_u}\exp\left(\frac{-f(\mathbf{r}(t))}{s_u}\right) & \text{if } f(\mathbf{r}(t)) \geq 0 \\
        \frac{1}{s_u}\left(1 - \frac{1}{2}\exp\left(\frac{f(\mathbf{r}(t))}{s_u}\right)\right) & \text{otherwise}
        \end{cases}.
    \end{aligned}
\end{equation}

In the refinement stage, we adopt the unbiased TUVR formulation~\cite{zhang2023towards} with the uncertainty-aware scale parameter $s_u$ as follows:

\begin{equation} 
\label{eq:supp_2}
    \begin{aligned}
        \sigma(\mathbf{r}(t)) &= \Phi_{s_u}^{\text{TUVR}}(f(\mathbf{r}(t))) \\
        =&
        \begin{cases}
        \frac{1}{s_u}\exp\left(\frac{-f(\mathbf{r}(t))}{s_u|f'(\mathbf{r}(t))|}\right) & \text{if } f(\mathbf{r}(t)) \geq 0 \\
        \frac{2}{s_u}\left(1 - \frac{1}{2}\exp\left(\frac{f(\mathbf{r}(t))}{s_u|f'(\mathbf{r}(t))|}\right)\right) & \text{otherwise}
        \end{cases}.
    \end{aligned}
\end{equation}


\subsubsection{Adaptive Regularization Weight.}
To compute the adaptive regularization weight, we normalize the predicted SDF uncertainty within each batch. The normalized uncertainty $\tilde{\beta}_f^2(\mathbf{r}(t))$ is defined as:

\begin{equation} 
\label{eq:supp_3}
    \tilde{\beta}_f^2(\mathbf{r}(t)) = \frac{\beta_f^2(\mathbf{r}(t)) - \beta_f^2(\mathbf{r}(t))_{\min}}{\beta_f^2(\mathbf{r}(t))_{\max} - \beta_f^2(\mathbf{r}(t))_{\min} + \varepsilon},
\end{equation}
where the min and max are computed over all sampled points in the current batch, and $\varepsilon$ is a small constant added for numerical stability.


\subsubsection{Stochastic-Step Numerical Gradient.}
We adopt the stochastic-step numerical gradient estimation technique proposed in NeuRodin~\cite{wang2024neurodin}. Specifically, the $x$-component of the estimated gradient $\hat{\nabla} f$ is computed as follows:

\begin{equation}
\label{eq:supp_4}
    \hat{\nabla}_x f(\mathbf{r}(t)) = \frac{f\left(\mathbf{r}(t)+\boldsymbol{\epsilon}_x\right)-f\left(\mathbf{r}(t)-\boldsymbol{\epsilon}_x\right)}{2\epsilon_x},
\end{equation}
where $\boldsymbol{\epsilon}_x = (\epsilon_x, 0, 0)$ and $\epsilon_x$ is sampled from a uniform distribution $\epsilon_x \sim U(0, \epsilon_{\text{max}})$.


\subsubsection{Additional Losses.}
We incorporate explicit bias correction loss $\mathcal{L}_{\text{bias}}$ proposed by NeuRodin~\cite{wang2024neurodin} during the initial stage of training. It is defined as follows:

\begin{equation}
\label{eq:supp_5}
    \begin{aligned}
        \mathcal{L}_{\text{bias}} &= \frac{1}{m}\sum_{\mathbf{r}\in \mathcal{R}} \max\left(f(\mathbf{r}(t^* + \epsilon_{\text{bias}})), 0\right), \\
        t^* &= \underset{t \in (0, +\infty)}{\arg\max} \ T(\mathbf{r}(t))\alpha(\mathbf{r}(t)),
    \end{aligned}
\end{equation}
where $\epsilon_{\text{bias}}$ is a small offset set to $0.0005$.

We additionally incorporate the color regularization loss proposed by PermutoSDF~\cite{rosu2023permutosdf} during the refinement stage of training. Given an MLP layer defined as $y=\sigma_\text{act}(W_ix+b_i)$ and a trainable Lipschitz bound $\kappa_i$ for that layer, the weight matrix $W_i$ is replaced with a normalized version $\widehat{W}_i$ as follows:

\begin{equation}
\label{eq:supp_6}
    \widehat{W}_i = \textit{m}\,(W_i, \text{softplus}(\kappa_i)),
\end{equation}
where $\text{softplus}(\kappa_i) = \ln(1 + e^{\kappa_i})$, and the normalization function $\textit{m}(\cdot)$ rescales each row of $W_i$ such that the absolute row sum does not exceed $\text{softplus}(\kappa_i)$. The color regularization loss is then defined as follows:

\begin{equation}
\label{eq:supp_7}
    \mathcal{L}_\text{Lipschitz}= \prod_{l}^{i=1} \text{softplus}(\kappa_i).
\end{equation}

\subsection{Hyperparameters}

We present the additional hyperparameters used in NeuDonatello. The spatial offset $\epsilon$ used for the smoothness constraint $\mathcal{L}_{\text{smooth}}$ is set to $0.01$ along the tangent direction. The weighting factor $\gamma$ in $\mathcal{L}_{\text{NLL}}$ is set to $0.5$.

\subsection{Evaluation Details}

For the ScanNet++ dataset, we evaluate reconstruction quality using six metrics (Accuracy, Completion, Precision, Recall, Chamfer Distance, and F1-score (threshold: $0.025$)) following~\cite{wang2022neuris, yu2022monosdf}, computed between the predicted mesh and the reference mesh generated from laser-scanned point clouds. For the Tanks and Temples dataset, we follow the official evaluation protocol and compute metrics on the training subset using the dataset's provided Python evaluation toolkit.\footnote{\url{https://github.com/isl-org/TanksAndTemples/tree/master/python_toolbox/evaluation}}

\section{Additional Analysis}

\subsection{Analysis on Uncertainty Estimation}

\begin{figure}[t!]
    \centering
        \centerline{\includegraphics[width=0.8\columnwidth]{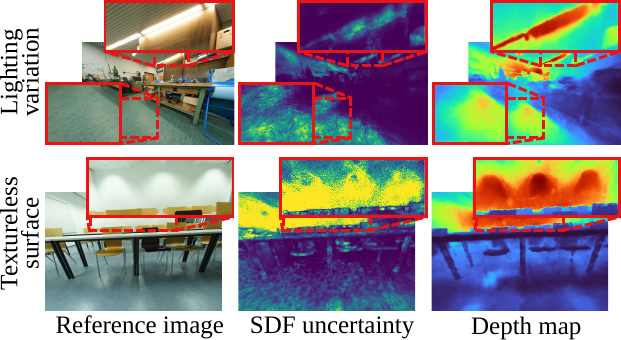}}
        \caption{Analysis on uncertainty estimation. We visualize estimated SDF uncertainty with produced depth map at the initial stage.}
        \label{fig:uncert_analysis_a}
\end{figure}

To analyze the effectiveness of our uncertainty estimation, we visualize the predicted SDF uncertainty alongside rendered depth maps in \cref{fig:uncert_analysis_a}. We observe a clear correspondence: regions with high SDF uncertainty align with areas where depth predictions are inaccurate or inconsistent. In the upper row, lighting variations introduce geometric ambiguity, resulting in erroneous depth prediction. Our uncertainty model correctly highlights these regions. In the bottom row, textureless surfaces make accurate reconstruction difficult, leading to imprecise depth prediction. Again, our uncertainty model successfully identifies these areas. These results demonstrate that our model successfully identifies geometrically ambiguous regions, such as occlusions, textureless surfaces, or lighting variations, where reliable reconstruction is inherently difficult. The learned SDF uncertainty provides a meaningful signal, accurately reflecting the confidence of the model in its geometric predictions and guiding the downstream modules accordingly.

\begin{figure}[t!]
    \centering
        \centerline{\includegraphics[width=0.8\columnwidth]{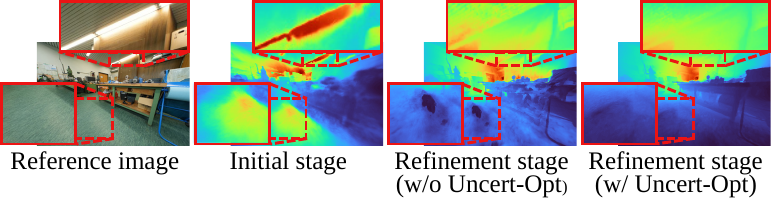}}
        \caption{Depth map at the initial stage and refinement stage with and without our uncertainty-aware optimization.}
        \label{fig:uncert_analysis_b}
\end{figure}


In \cref{fig:uncert_analysis_b}, we compare depth maps in regions with lighting variations, with and without our uncertainty-aware optimization. Without the uncertainty-aware modules, the network struggles to learn accurate geometry, resulting in distorted or inconsistent depth predictions. In contrast, when guided by uncertainty-aware optimization, the network identifies ambiguous regions and adaptively adjusts the optimization process, enabling accurate geometry reconstruction and improved depth consistency. These results further confirm the effectiveness of our uncertainty-aware modules.


\subsection{Analysis on Monte Carlo Sampling}

To further evaluate the impact of our uncertainty modeling, we conduct an ablation study by varying the number of Monte Carlo samples $N_{\text{mc}}$. We experiment with $N_{\text{mc}} = 3$, $5$, $10$, and $20$ on the ScanNet++ dataset and report the mean reconstruction quality across eight representative scenes in \cref{tab:mc_smaple_ablation}. The results show that increasing $N_{\text{mc}}$ generally improves reconstruction quality, with performance peaking at $N_{\text{mc}} = 10$. However, increasing the number of Monte Carlo samples beyond a moderate range yields negligible gains. For example, using too many samples (e.g., $N_{\text{mc}} = 20$) does not noticeably improve reconstruction quality but instead increases computational complexity, slows optimization, and introduces additional variance into the training signal.

\begin{figure*}[t]
    \centering
        \centerline{\includegraphics[width=\linewidth]{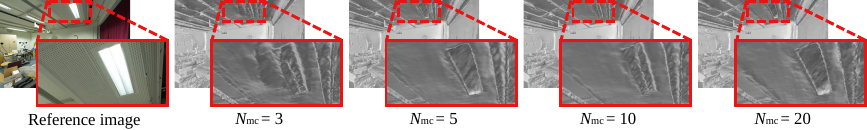}}
        \caption{Reconstructed meshes using different numbers of Monte Carlo samples.}
        \label{fig:1}
\end{figure*}

We also visualize the reconstructed mesh and the uncertainty maps with different numbers of Monte Carlo samples in \cref{fig:1} and \cref{fig:2}. With only $3$ samples, the uncertainty network struggles to localize ambiguous regions, producing noisy estimates and capturing only a subset of the uncertain areas. As the number of samples increases, the network improves in identifying regions of geometric ambiguity with greater accuracy and consistency. Based on these observations, we choose $N_{\text{mc}} = 10$, as it provides reliable uncertainty estimation while avoiding the unnecessary computational overhead.

\setlength\tabcolsep{1.3em}
\begin{table}[t]
    \centering
    \begin{tabularx}{\columnwidth}{c|cccccc}
        \specialrule{.2em}{.1em}{.1em} 
        $N_{\text{mc}}$
        & \makecell{Acc. \\ ($\downarrow$)}
        & \makecell{Comp. \\ ($\downarrow$)}
        & \makecell{Pre. \\ ($\uparrow$)}
        & \makecell{Recall \\ ($\uparrow$)}
        & \makecell{Chamfer \\ ($\downarrow$)}
        & \makecell{F1-Score \\ ($\uparrow$)} \\
        \cmidrule(l{.33em}r{.33em}){1-7}
        3 & 0.082 & 0.049 & 0.582 & 0.608 & 0.066 & 0.593 \\
        5 & 0.081 & 0.048 & 0.593 & 0.620 & 0.065 & 0.604 \\
        10 & \bfseries{0.052} & \bfseries{0.047} & \bfseries{0.618} & 0.627 & \bfseries{0.049} & \bfseries{0.621} \\
        20 & 0.062 & \bfseries{0.047} & 0.611 & \bfseries{0.628} & 0.055 & 0.620 
        \\
        \specialrule{.2em}{.1em}{.1em}
    \end{tabularx}%
    \caption{
        Ablation on the number of Monte Carlo samples.
    }
    \label{tab:mc_smaple_ablation}
\end{table}

\begin{figure*}[t!]
    \centering
        \centerline{\includegraphics[width=\linewidth]{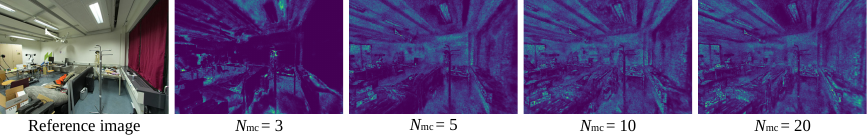}}
        \caption{Visualization of SDF uncertainty maps using different numbers of Monte Carlo samples.}
        \label{fig:2}
\end{figure*}

\subsection{Analysis on Uncertainty-Aware Adaptive Geometric Regularization}

We evaluate the effect of applying uncertainty-aware adaptive geometric regularization in the initial stage and report the performance in the ScanNet++ dataset in \cref{tab:adaptive_reg_ablation}. The results show a performance drop when the regularization is applied in both the initial and refinement stages, compared with applying it only in the refinement stage. We attribute this to over-regularization early in training. During the initial stage, the geometry is still noisy and the predicted normals are unreliable. Applying strong regularization, especially in ambiguous regions where uncertainty is high, can distort the surface geometry. This issue is compounded by the fact that uncertainty estimates are not yet accurate in the early stage. In contrast, applying adaptive regularization only after the initial stage, when the geometry has stabilized and uncertainty estimates are more reliable, leads to improved reconstruction performance. A visual comparison illustrating this effect is provided in \cref{fig:3}.


\subsection{Analysis on Uncertainty-Aware SDF-to-Density Conversion}

We demonstrate the impact of our uncertainty-aware SDF-to-density conversion by visualizing the predicted scale parameter on the Tanks and Temples dataset in \cref{fig:4}. The scale value should be low in regions with strong photometric signals, enabling precise surface reconstruction. In contrast, it should be high in ambiguous areas to avoid converging to incorrect surfaces. As shown in \cref{fig:4}, this behavior is clearly observed: geometrically certain areas such as edges and texture-rich regions exhibit low scale values, while ambiguous regions maintain higher scale values, preventing inaccurate surface reconstruction.


\setlength\tabcolsep{0.77em}
\begin{table}[t!]
    \centering
    \begin{tabularx}{\columnwidth}{cc|cccccc}
        \specialrule{.2em}{.1em}{.1em}
        \makecell{Initial \\ stage} & \makecell{Refinement \\ stage}
        & \makecell{Acc. \\ ($\downarrow$)}
        & \makecell{Comp. \\ ($\downarrow$)}
        & \makecell{Pre. \\ ($\uparrow$)}
        & \makecell{Recall \\ ($\uparrow$)}
        & \makecell{Chamfer \\ ($\downarrow$)}
        & \makecell{F1-Score \\ ($\uparrow$)} \\
        \cmidrule(l{0.5em}r{.5em}){1-8}
        & & 0.081 & 0.049 & 0.583 & 0.603 & 0.065 & 0.593 \\
        \checkmark & \checkmark & 0.084 & \bfseries{0.047} & 0.595 & \bfseries{0.627} & 0.065 & 0.611 \\
        & \checkmark & \bfseries{0.052} & \bfseries{0.047} &\bfseries{ 0.618} & \bfseries{0.627} & \bfseries{0.049} & \bfseries{0.621} \\
        \specialrule{.2em}{.1em}{.1em}
    \end{tabularx}
    \caption{
        Ablation on uncertainty-aware adaptive geometric regularization.
    }
    \label{tab:adaptive_reg_ablation}
\end{table}

\begin{figure}[t!]
    \centering
        \centerline{\includegraphics[width=0.7\columnwidth]{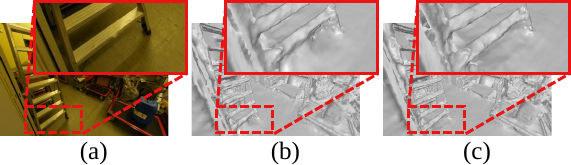}}
        \caption{(a) Region with complex geometry. (b) Applying adaptive geometric regularization in both the initial and refinement stages oversmooths the structure. (c) Applying it only during the refinement stage better preserves details and leads to more accurate reconstruction.}
        \label{fig:3}
\end{figure}

\begin{figure}[t!]
    \centering
        \centerline{\includegraphics[width=0.7\columnwidth]{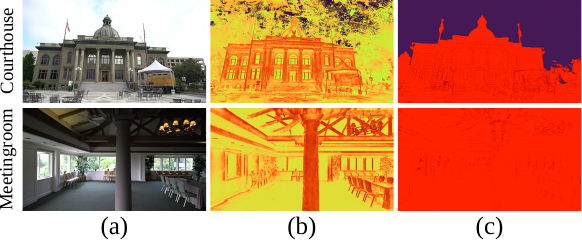}}
        \caption{(a) Reference image. (b) Scale parameter after the initial stage: ambiguous regions exhibit high values (bright), while geometrically reliable regions show low values (dark). (c) Scale parameter after the refinement stage: with upper bound scheduling, all regions converge to low values, accurately localizing the surface.}
        \label{fig:4}
\end{figure}

While effective, this formulation can limit the model’s ability to converge to the surface in ambiguous regions. To address this, we manually control the upper bound of the scale parameter. Specifically, in the initial stage, the upper bound decreases exponentially from infinity to $0.01$. In the refinement stage, the upper bound is further reduced by decreasing exponentially from infinity to $3 \times 10^{-4}$.

\section{Additional Results}

\subsection{Comparison with Explicit Methods}

We compare NeuDonatello against recent explicit representations based on 3D Gaussian Splatting (3DGS)~\cite{kerbl20233d}. Specifically, we compare with mesh reconstruction approaches such as SuGaR~\cite{guedon2024sugar}, 2DGS~\cite{huang20242d}, and GOF~\cite{yu2024gaussian}.

On the ScanNet++ dataset, NeuDonatello achieves the highest performance on every metric, clearly outperforming the strongest 3DGS-based baselines, as summarized in \cref{tab:scannetpp_gs}. The qualitative comparison in \cref{fig:5} further highlights that our method produces cleaner surfaces with fewer artifacts while better preserving geometric details.

On the Tanks and Temples dataset, NeuDonatello maintains this advantage, achieving the top F1-score across all methods, as shown in \cref{tab:tnt_gs}. These results demonstrate the robustness of our approach in reconstructing accurate geometry, even in complex environments.

\setlength\tabcolsep{0.75em}
\begin{table*}[t]
    \centering
    \begin{tabularx}{\textwidth}{@{}lrcccccc}
        \specialrule{.2em}{.1em}{.1em} 
        & & \multicolumn{6}{c}{Metric} \\ \cmidrule(l{.15em}r{.15em}){3-8}
        & \multicolumn{1}{c}{Method} 
        & \makecell{Acc. \\ ($\downarrow$)} 
        & \makecell{Comp. \\ ($\downarrow$)} 
        & \makecell{Pre. \\ ($\uparrow$)}
        & \makecell{Recall \\ ($\uparrow$)}
        & \makecell{Chamfer \\ ($\downarrow$)}
        & \makecell{F1-score \\ ($\uparrow$)} \\
        \cmidrule(l{0em}r{.15em}){2-2}
        \cmidrule(l{.15em}r{.15em}){3-8}
        & SuGaR~\cite{guedon2024sugar} & 0.059 & 0.061 & 0.487 & 0.411 & 0.060 & 0.411  \\
        & 2DGS~\cite{huang20242d} & 0.082 & 0.053 & 0.489 & 0.532 & 0.068 & 0.509 \\
        & GOF~\cite{yu2024gaussian} & \underline{0.057} & \bfseries{0.047} & \underline{0.570} & \underline{0.536} & \underline{0.052} & \underline{0.548} \\
        & NeuDonatello (Ours) & \bfseries{0.052} & \bfseries{0.047} & \bfseries{0.618} & \bfseries{0.627} & \bfseries{0.049} & \bfseries{0.621} \\
        \specialrule{.2em}{.1em}{.1em}
    \end{tabularx}%
    \caption{
        Quantitative results compared with explicit methods on ScanNet++ dataset.
        Best result (bold). Second best result (underlined).
    }
    \label{tab:scannetpp_gs}
\end{table*}

\begin{figure*}[t]
    \centering
        \centerline{\includegraphics[width=\linewidth]{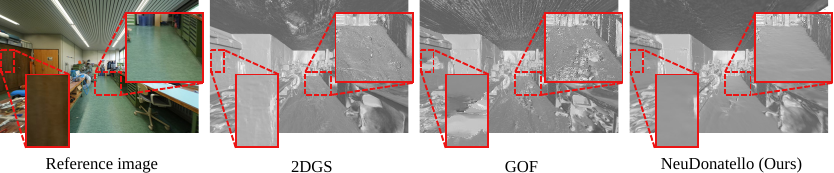}}
        \caption{Qualitative results compared with explicit methods on ScanNet++ dataset.}
        \label{fig:5}
\end{figure*}


\setlength\tabcolsep{0.22em}
\begin{table*}[t!]
    \renewcommand{\arraystretch}{1.0}
    \centering
    \begin{tabularx}{\textwidth}{@{}lrccccccc}
        \specialrule{.2em}{.1em}{.1em}
        & & \multicolumn{6}{c}{Scene} \\ \cmidrule(l{.2em}r{.2em}){3-8}
        & \multicolumn{1}{c}{Method} & Barn & Caterpillar & Courthouse & Ignatius & Meetingroom & Truck & Mean \\
        \cmidrule(l{0em}r{.2em}){2-2}
        \cmidrule(l{.2em}r{.2em}){3-8}
        \cmidrule(l{.2em}r{0.2em}){9-9}
        & SuGaR~\cite{guedon2024sugar} & 0.14 & 0.16 & 0.08 & 0.33 & 0.15 & 0.26 & 0.19 \\
        & 2DGS~\cite{huang20242d} & 0.42 & 0.23 & 0.16 & 0.51 & 0.17 & 0.45 & 0.32 \\
        & GOF~\cite{yu2024gaussian} & \underline{0.51} & \bfseries{0.41} & \bfseries{0.28} & \underline{0.68} & \underline{0.28} & \underline{0.59} & \underline{0.46} \\
        & NeuDonatello (Ours) & \bfseries{0.71} & \bfseries{0.37} & \underline{0.22} & \bfseries{0.85} & \bfseries{0.44} & \bfseries{0.48} & \bfseries{0.51} \\
        \specialrule{.2em}{.1em}{.1em}
    \end{tabularx}%
    \caption{
       Quantitative results (F1-score ($\uparrow$)) compared with explicit methods on Tanks and Temples dataset. Best result (bold). Second best result (underlined)
    }
    \label{tab:tnt_gs}
\end{table*}


\subsection{Comparison with Uncertainty-Aware Methods}

\begin{figure*}[t!]
    \centering
        \centerline{\includegraphics[width=1.0\linewidth]{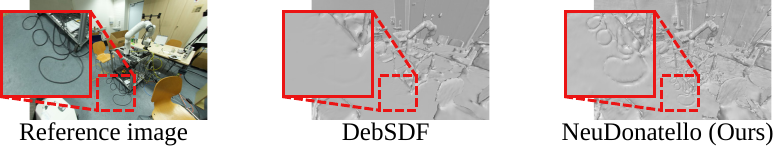}}
        \caption{Qualitative comparison against DebSDF on ScanNet++ dataset.}
        \label{fig:depsdf_compare}
\end{figure*}

We compare NeuDonatello against DebSDF~\cite{xiao2024debsdf}, a recent method that incorporates uncertainty estimation over depth and normal priors for neural surface reconstruction. Although DebSDF models uncertainty to filter unreliable priors, it still oversmooths fine geometric details, as shown in \cref{fig:depsdf_compare}. In contrast, NeuDonatello preserves thin structures such as cables and leverages uncertainty as an intrinsic property of the RGB-only inverse rendering process to accurately reconstruct geometry in ambiguous regions without relying on external priors. NeuDonatello also achieves competitive quantitative performance without explicit priors, as summarized in \cref{tab:scannetpp_debsdf}.


\setlength\tabcolsep{0.7em}
\begin{table*}[t]
    \centering
    \begin{tabularx}{\textwidth}{@{}lrcccccc}
        \specialrule{.2em}{.1em}{.1em} 
        & & \multicolumn{6}{c}{Metric} \\ \cmidrule(l{.15em}r{.15em}){3-8}
        & \multicolumn{1}{c}{Method} 
        & \makecell{Acc. \\ ($\downarrow$)} 
        & \makecell{Comp. \\ ($\downarrow$)} 
        & \makecell{Pre. \\ ($\uparrow$)}
        & \makecell{Recall \\ ($\uparrow$)}
        & \makecell{Chamfer \\ ($\downarrow$)}
        & \makecell{F1-score \\ ($\uparrow$)} \\
        \cmidrule(l{0em}r{.15em}){2-2}
        \cmidrule(l{.15em}r{.15em}){3-8}
        & DebSDF~\cite{xiao2024debsdf} & \bfseries{0.048} & \bfseries{0.043} & 0.605 & 0.612 & \bfseries{0.045} & 0.609  \\
        & NeuDonatello (Ours) & 0.052 & 0.047 & \bfseries{0.618} & \bfseries{0.627} & 0.049 & \bfseries{0.621} \\
        \specialrule{.2em}{.1em}{.1em}
    \end{tabularx}%
    \caption{
        Quantitative results compared with DebSDF on ScanNet++ dataset.
    }
    \label{tab:scannetpp_debsdf}
\end{table*}


\subsection{ScanNet++}

We present additional per-scene quantitative results on the ScanNet++ dataset in \cref{tab:additional_metric_scannet++_detail} and qualitative results in \cref{fig:6} and \cref{fig:7}. NeuDonatello consistently outperforms other baselines, demonstrating visibly improved reconstructions, particularly in geometrically ambiguous regions such as textureless surfaces, lighting variations, and occlusions.

\subsection{Tanks and Temples}

\begin{figure*}[t]
    \centering
        \centerline{\includegraphics[width=\linewidth]{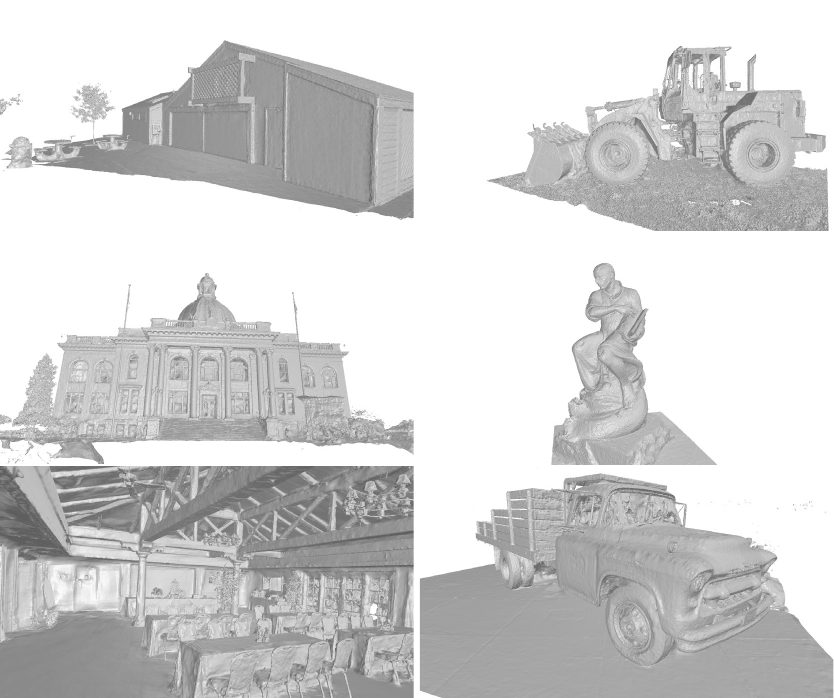}}
        \caption{Additional qualitative results on the Tanks and Temples dataset.}
        \label{fig:8}
\end{figure*}

We show additional qualitative results on the Tanks and Temples dataset in \cref{fig:8}. NeuDonatello demonstrates visually compelling reconstructions in both indoor and outdoor scenes. Our method accurately captures fine-grained geometric details and maintains surface fidelity, even in challenging regions.


\section{Limitation}

Despite its effectiveness, NeuDonatello has several limitations. First, the final quality of the reconstruction is influenced by the initial stage. If the early geometry estimation falls into severely incorrect surfaces, the refinement stage may not fully recover accurate geometry. Second, although Monte Carlo sampling provides a simple and effective way of estimating uncertainty, it increases the computational cost during training. Lastly, our model is sensitive to hyperparameters, as it directly modifies the weighting of loss terms based on uncertainty. This requires careful tuning to ensure stable and effective training.


\section{Societal Impact}

NeuDonatello offers benefits for fields such as architecture, virtual reality, and robotics by enabling high-fidelity 3D reconstruction from casually captured RGB images. In architecture, it can assist in creating accurate digital twins of indoor spaces; in virtual and augmented reality, it enables immersive scene capture; and in robotics, it supports scene understanding for navigation and manipulation. However, like many vision-based systems, it raises potential privacy concerns when used to reconstruct real-world environments without consent. Its computational cost may also limit broader accessibility. Overall, while NeuDonatello offers promising advances in 3D perception, careful consideration is needed regarding its deployment and societal implications.


\setlength\tabcolsep{0.8em}
\begin{table*}[t!]
    \renewcommand{\arraystretch}{1.2}
    \centering
    \scriptsize
    \begin{tabularx}{\linewidth}{c|r|cc|cccc}
    \toprule[1.5pt]
    Scene                       & \multicolumn{1}{c|}{Metric} & \makecell{MonoSDF \\ MLP\textsuperscript{*} \\ \cite{yu2022monosdf}}
    & \makecell{MonoSDF \\ Grid\textsuperscript{*} \\ \cite{yu2022monosdf}}
    & \makecell{VolSDF \\ \cite{yariv2021volume}}
    & \makecell{Neuralangelo \\ \cite{li2023neuralangelo}}
    & \makecell{NeuRodin \\ \cite{wang2024neurodin}}
    & \makecell{NeuDonatello \\ (Ours)}  \\ \midrule
    \multirow{6}{*}{\rotatebox[origin=c]{90}{\strut 0e75f3c4d9}}
    & Acc ($\downarrow$)
        & \underline{0.044}       
        & 0.063
        & 0.085
        & 0.300        
        & 0.136
        & \bfseries{0.033}  \\
    & Comp ($\downarrow$)
        & \underline{0.031}     
        & \bfseries{0.028}                
        & 0.112  
        & 0.176        
        & 0.048                 
        & 0.039 \\
    & Pre ($\uparrow$)            
        & \underline{0.628}       
        & 0.596                           
        & 0.337  
        & 0.285        
        & 0.580                 
        & \bfseries{0.688} \\
    & Recall ($\uparrow$)         
        & \underline{0.721}       
        & 0.715                           
        & 0.307  
        & 0.397        
        & 0.691                 
        & \bfseries{0.749} \\
    & Chamfer ($\downarrow$)      
        & \underline{0.038}       
        & 0.046                           
        & 0.099  
        & 0.238        
        & 0.092                 
        & \bfseries{0.036} \\
    & F1-score ($\uparrow$)       
        & \underline{0.671}       
        & 0.650                           
        & 0.321  
        & 0.332        
        & 0.626                 
        & \bfseries{0.717} \\
    \midrule
    \multirow{6}{*}{\rotatebox[origin=c]{90}{\strut 036bce3393}}
    & Acc ($\downarrow$)          
        & 0.052       
        & 0.041                           
        & 0.078  
        & 0.085        
        & \underline{0.031}     
        & \bfseries{0.030} \\
    & Comp ($\downarrow$)         
        & 0.074     
        & 0.045                           
        & 0.175  
        & 0.047        
        & \underline{0.034}     
        & \bfseries{0.032} \\
    & Pre ($\uparrow$)            
        & 0.459    
        & 0.576                           
        & 0.418  
        & 0.572        
        & \bfseries{0.674}      
        & \underline{0.673} \\
    & Recall ($\uparrow$)         
        & 0.433       
        & 0.604                           
        & 0.317  
        & 0.637        
        & \underline{0.681}     
        & \bfseries{0.697} \\
    & Chamfer ($\downarrow$)      
        & 0.063       
        & 0.043                           
        & 0.127  
        & 0.066        
        & \underline{0.033}     
        & \bfseries{0.031} \\
    & F1-score ($\uparrow$)       
        & 0.445       
        & 0.590                           
        & 0.360  
        & 0.603        
        & \underline{0.677}     
        & \bfseries{0.685} \\
    \midrule
    \multirow{6}{*}{\rotatebox[origin=c]{90}{\strut 108ec0b806}}
    & Acc ($\downarrow$)          
        & 0.045       
        & 0.042                           
        & 0.141  
        & 0.066        
        & \underline{0.037}     
        & \bfseries{0.036} \\
    & Comp ($\downarrow$)         
        & 0.068     
        & \bfseries{0.046}                
        & 0.295  
        & 0.062        
        & 0.056                 
        & \underline{0.051} \\
    & Pre ($\uparrow$)            
        & \bfseries{0.622}            
        & 0.463                           
        & 0.529  
        & 0.331  
        & 0.565                 
        & \underline{0.608} \\
    & Recall ($\uparrow$)         
        & 0.411       
        & 0.531                           
        & 0.211  
        & \underline{0.561}    
        & 0.552                 
        & \bfseries{0.584} \\
    & Chamfer ($\downarrow$)      
        & 0.057       
        & \bfseries{0.044}                
        & 0.218  
        & 0.064        
        & \underline{0.047}     
        & \bfseries{0.044}  \\
    & F1-score ($\uparrow$)       
        & 0.426       
        & 0.530                           
        & 0.258  
        & 0.563        
        & \underline{0.578}     
        & \bfseries{0.602} \\
    \midrule
    \multirow{6}{*}{\rotatebox[origin=c]{90}{\strut 21d970d8de}}
    & Acc ($\downarrow$)          
        & \underline{0.051}       
        & \bfseries{0.046}                
        & 0.079  
        & 0.112        
        & 0.078                 
        & 0.080 \\
    & Comp ($\downarrow$)         
        & 0.054     
        & \underline{0.036}               
        & 0.142  
        & 0.127        
        & \bfseries{0.035}      
        & 0.037 \\
    & Pre ($\uparrow$)            
        & 0.393    
        & 0.482                           
        & 0.344  
        & 0.467        
        & \bfseries{0.574}      
        & \underline{0.551} \\
    & Recall ($\uparrow$)         
        & 0.422    
        & 0.554                           
        & 0.294  
        & 0.449        
        & \bfseries{0.677}      
        & \underline{0.657} \\
    & Chamfer ($\downarrow$)      
        & \underline{0.053}       
        & \bfseries{0.041}                
        & 0.111  
        & 0.120        
        & 0.057                 
        & 0.059 \\
    & F1-score ($\uparrow$)       
        & 0.407    
        & 0.515                           
        & 0.317  
        & 0.458        
        & \bfseries{0.621}      
        & \underline{0.600} \\
    \midrule
    \multirow{6}{*}{\rotatebox[origin=c]{90}{\strut 355e5e32db}}
    & Acc ($\downarrow$)          
        & 0.036       
        & \underline{0.034}               
        & 0.070  
        & 0.047        
        & 0.048                 
        & \bfseries{0.032} \\
    & Comp ($\downarrow$)         
        & \underline{0.047}     
        & \bfseries{0.037}                
        & 0.130  
        & 0.048        
        & 0.058                
        & 0.057 \\
    & Pre ($\uparrow$)            
        & 0.524       
        & 0.582                           
        & 0.415  
        & \bfseries{0.683}     
        & 0.642                 
        & \underline{0.667} \\
    & Recall ($\uparrow$)         
        & 0.515       
        & 0.596                           
        & 0.323  
        & \bfseries{0.665}     
        & 0.601                 
        & \underline{0.610} \\
    & Chamfer ($\downarrow$)      
        & \underline{0.042}       
        & \bfseries{0.036}                
        & 0.100  
        & 0.048        
        & 0.053                 
        & 0.045 \\
    & F1-score ($\uparrow$)       
        & 0.519       
        & 0.589                           
        & 0.363  
        & \bfseries{0.674}     
        & 0.623                 
        & \underline{0.637} \\
    \midrule
    \multirow{6}{*}{\rotatebox[origin=c]{90}{\strut 578511c8a9}}
    & Acc ($\downarrow$)          
        & \underline{0.045}       
        & \bfseries{0.042}                
        & 0.197  
        & 0.237        
        & 0.135                 
        & 0.067 \\
    & Comp ($\downarrow$)         
        & \underline{0.049}     
        & \bfseries{0.037}                
        & 0.222  
        & 0.081        
        & 0.063                 
        & 0.052 \\
    & Pre ($\uparrow$)            
        & 0.548       
        & \bfseries{0.569}                
        & 0.235  
        & 0.410        
        & 0.508                 
        & \underline{0.561} \\
    & Recall ($\uparrow$)         
        & 0.571       
        & \bfseries{0.623}                
        & 0.198  
        & 0.532        
        & 0.552                 
        & \underline{0.612} \\
    & Chamfer ($\downarrow$)      
        & \underline{0.047}       
        & \bfseries{0.040}                
        & 0.210  
        & 0.159        
        & 0.099                 
        & 0.060 \\
    & F1-score ($\uparrow$)       
        & 0.559       
        & \bfseries{0.595}                
        & 0.215  
        & 0.463        
        & 0.529                 
        & \underline{0.586} \\
    \midrule
    \multirow{6}{*}{\rotatebox[origin=c]{90}{\strut 7f4d173c9c}}
    & Acc ($\downarrow$)          
        & \underline{0.113}       
        & 0.176                           
        & 0.175  
        & 0.194        
        & 0.156                 
        & \bfseries{0.055} \\
    & Comp ($\downarrow$)         
        & \underline{0.025}     
        & \bfseries{0.022}                
        & 0.192  
        & 0.034        
        & 0.043                 
        & 0.043 \\
    & Pre ($\uparrow$)            
        & \bfseries{0.706}       
        & 0.684                           
        & 0.282  
        & 0.649        
        & 0.632                 
        & \underline{0.692} \\
    & Recall ($\uparrow$)         
        & \underline{0.759}       
        & \bfseries{0.797}                
        & 0.237  
        & 0.721        
        & 0.632                 
        & 0.644 \\
    & Chamfer ($\downarrow$)      
        & \underline{0.069}       
        & 0.099                           
        & 0.184  
        & 0.114        
        & 0.100                 
        & \bfseries{0.049} \\
    & F1-score ($\uparrow$)       
        & \underline{0.732}       
        & \bfseries{0.736}                
        & 0.257  
        & 0.683        
        & 0.632                 
        & 0.667 \\
    \midrule
    \multirow{6}{*}{\rotatebox[origin=c]{90}{\strut 09c1414f1b}}
    & Acc ($\downarrow$)          
        & \bfseries{0.038}    
        & 0.077                           
        & 0.124  
        & 0.206        
        & \underline{0.076}     
        & 0.084 \\
    & Comp ($\downarrow$)         
        & 0.071    
        & 0.065                           
        & 0.277  
        & 0.158        
        & \underline{0.063}     
        & \bfseries{0.061} \\
    & Pre ($\uparrow$)            
        & \bfseries{0.616}    
        & \bfseries{0.616}                
        & 0.325  
        & 0.504        
        & \underline{0.515}     
        & 0.493 \\
    & Recall ($\uparrow$)         
        & \underline{0.537}       
        & \bfseries{0.574}                
        & 0.245  
        & 0.493        
        & 0.459                 
        & 0.460 \\
    & Chamfer ($\downarrow$)      
        & \bfseries{0.055}    
        & 0.071                           
        & 0.201  
        & 0.182        
        & \underline{0.070}     
        & 0.073 \\
    & F1-score ($\uparrow$)       
        & \underline{0.574}       
        & \bfseries{0.594}                
        & 0.279  
        & 0.498        
        & 0.486                 
        & 0.476 \\
    \bottomrule[1.5pt]
    \end{tabularx}
    \caption{
        Detailed results for ScanNet++ benchmark. Best result (bold). Second best result (underlined). \textsuperscript{*} indicates methods that use monocular depth and normal priors.
    } 
    \label{tab:additional_metric_scannet++_detail}
\end{table*}


\begin{figure*}[t]
    \centering
        \centerline{\includegraphics[width=\linewidth]{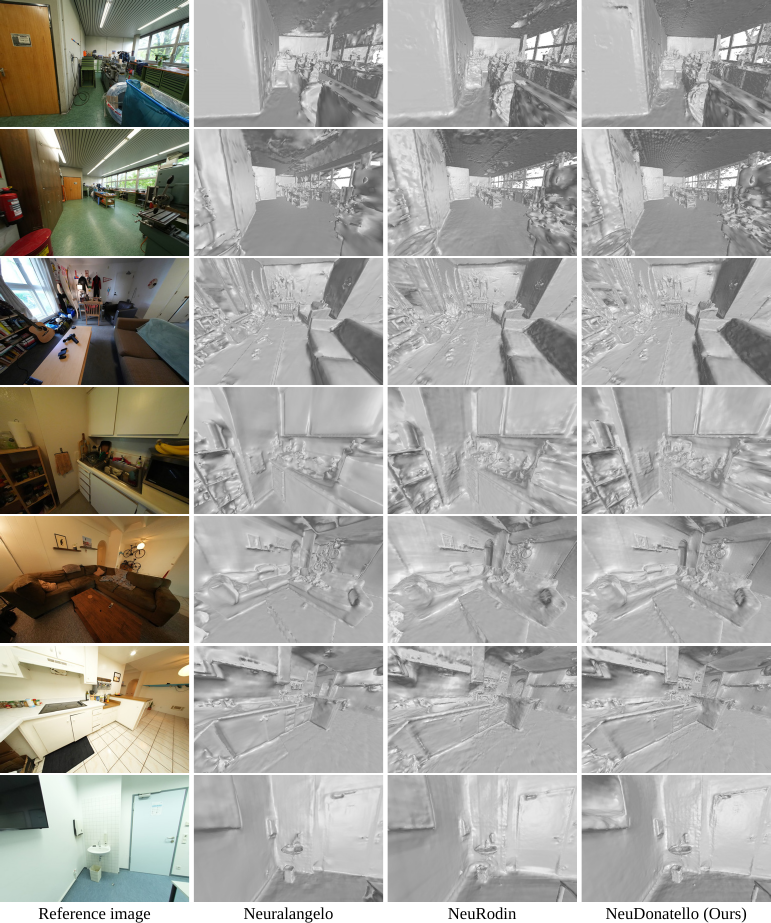}}
        \caption{Additional qualitative results on the ScanNet++ dataset.}
        \label{fig:6}
\end{figure*}

\begin{figure*}[t]
    \centering
        \centerline{\includegraphics[width=\linewidth]{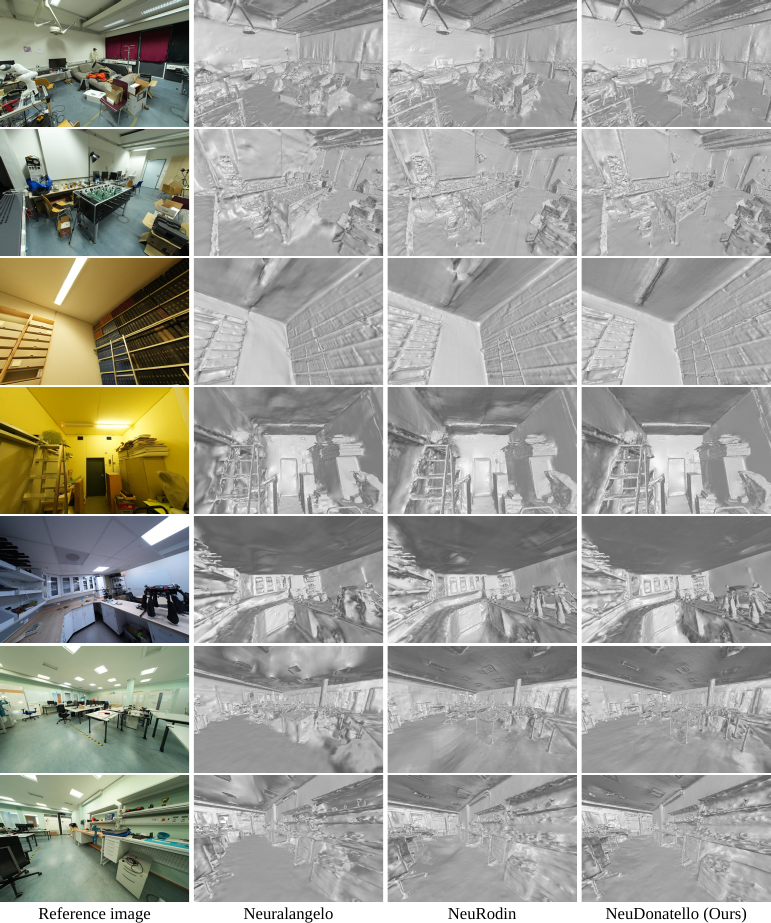}}
        \caption{Additional qualitative results on the ScanNet++ dataset.}
        \label{fig:7}
\end{figure*}

\end{document}